\documentclass[conference]{IEEEtran}
\IEEEoverridecommandlockouts
\usepackage{cite}
\usepackage{amsmath,amssymb,amsfonts}
\usepackage{algorithmic}
\usepackage{graphicx}
\usepackage{textcomp}
\usepackage{xcolor}
\usepackage{makecell}
\usepackage{csvsimple}
\usepackage{adjustbox}
\usepackage{hyperref}
\usepackage{float}
\usepackage{balance}
\usepackage{listings}
\usepackage{subcaption}
\usepackage{enumitem}
\def\BibTeX{{\rm B\kern-.05em{\sc i\kern-.025em b}\kern-.08em
    T\kern-.1667em\lower.7ex\hbox{E}\kern-.125emX}}

\def\tablename{Table}

\begin{document}

\title{MenTeR: A fully-automated \underline{M}ulti-ag\underline{enT} workflow for \underline{e}nd-to-end \underline{R}F/Analog Circuits Netlist Design\\
}


\author{\IEEEauthorblockN{Pin-Han Chen}
\IEEEauthorblockA{\textit{MediaTek Technology} \\
West Lafayette, Indiana \\
peter-ph.chen@mediatek.com}
\and
\IEEEauthorblockN{Yu-Sheng Lin}
\IEEEauthorblockA{\textit{MediaTek} \\
Hsinchu, Taiwan \\
bob-ys.lin@mediatek.com}
\and
\IEEEauthorblockN{Wei-Cheng Lee}
\IEEEauthorblockA{\textit{MediaTek} \\
Hsinchu, Taiwan \\
ds\_wei-cheng.lee@mediatek.com}
\and
\IEEEauthorblockN{Tin-Yu Leu}
\IEEEauthorblockA{\textit{MediaTek} \\
Hsinchu, Taiwan \\
magnus.leu@mediatek.com}
\and
\IEEEauthorblockN{Po-Hsiang Hsu}
\IEEEauthorblockA{\textit{MediaTek} \\
Hsinchu, Taiwan \\
stanley-ph.hsu@mediatek.com}
\and
\IEEEauthorblockN{Anjana Dissanayake}
\IEEEauthorblockA{\textit{MediaTek Technology} \\
West Lafayette, Indiana \\
anjana.dissanayake@mediatek.com}
\and
\IEEEauthorblockN{Sungjin Oh}
\IEEEauthorblockA{\textit{MediaTek Technology} \\
West Lafayette, Indiana \\
sungjin.oh@mediatek.com}
\and
\IEEEauthorblockN{Chinq-Shiun Chiu}
\IEEEauthorblockA{\textit{MediaTek USA} \\
West Lafayette, Indiana \\
cs.chiu@mediatek.com}
}

\maketitle
\begin{abstract}

RF/Analog design is essential for bridging digital technologies with real-world signals, ensuring the functionality and reliability of a wide range of electronic systems. However, analog design procedures are often intricate, time-consuming and reliant on expert intuition, and hinder the time and cost efficiency of circuit development. To overcome the limitations of the manual circuit design, we introduce MenTeR -- a multi-agent workflow integrated into an end-to-end analog design framework. By employing multiple specialized AI agents that collaboratively address different aspects of the design process, such as specification understanding, circuit optimization, and test bench validation, MenTeR reduces the dependency on frequent trial-and-error-style intervention. MenTeR not only accelerates the design cycle time but also facilitates a broader exploration of the design space, demonstrating robust capabilities in handling real-world analog systems. We believe that MenTeR lays the groundwork for future "RF/Analog Copilots" that can collaborate seamlessly with human designers.

\end{abstract}


\section{Introduction}
The recent progress of Large Language Models (LLMs) has led to an increasing numbers of LLM applications in scientific and engineering fields such as mathematical reasoning, pharmaceutical development, and chip design. For instance, in the field of digital circuit design, Liu et al. \cite{b1} introduced the first domain-adapted LLM, which demonstrated the potential of using legacy chip design documents to increase the design capabilities of LLM. BlockLove et al. \cite{b2} investigated the applications of LLMs in translating natural languages into Hardware Description Languages (HDL), showing the design example of an 8-bit microprocessor. In the field of RF/Analog circuit design, Lai et al. \cite{b3} proposed the first training-free LLM application to automate design of elementary circuit blocks through Python code generation. Liu et al. \cite{b4} introduced another LLM-based multi-agent system that automates multi-stage amplifier schematic design by integrating literature analysis, mathematical reasoning, and device sizing agents. 


While these works have established a good foundation for LLM applications in circuit design, there are still significant challenges for practical industrial deployment. For example, \cite{b3} leveraged LLMs to solve simplified analog design tasks and built a corresponding benchmark; however, there still exist several gaps between the benchmark problem sets of \cite{b3} and industrial-scale design problems. 
While \cite{b4} demonstrated that a multi-stage amplifier can be designed with a multi-agent system, the scalability of the approach to more complex amplifier designs or different circuit architectures is not yet explored.

To mitigate these gaps, we aim to build an AI-based solution for RF/Analog design, with the ultimate goal of assisted system-to-transistor-level implementation of complex circuits and systems such as Bandgap Reference (BGR) and Phase-Locked Loop (PLL). To achieve this goal, there are several questions that should be answered:
\begin{itemize}
\item What are the capabilities and limitations of current LLMs in RF/Analog circuits design? What techniques and data can we use to improve LLMs in reasoning required for practical designs?
\item RF/Analog circuits design is an intricate process that relies on the designer's intuitions and years of experience; how can we decompose this into multiple stages and integrate into an automated design flow?
\item How can we ensure the scalability and reliability of LLM-based systems for practical RF/Analog circuit designs? 
\end{itemize}

\begin{figure*}
    \includegraphics[width=\textwidth,height=12cm]{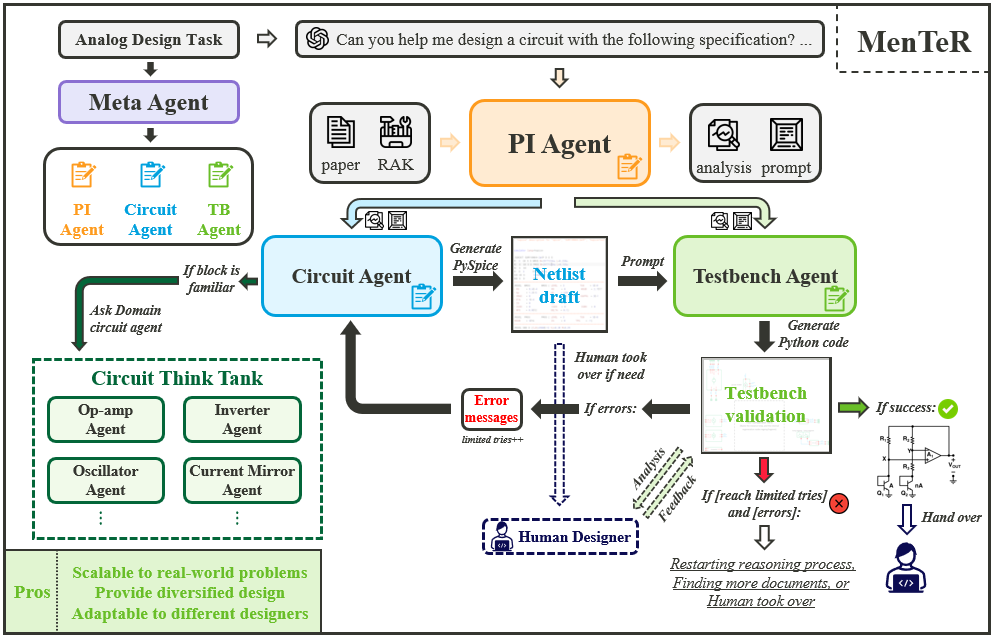}
    \vspace{-16pt}
    \caption{\textbf{Overview of MenTeR}. A fully-automated multi-agent workflow for RF/Analog circuit design that translates specifications to netlist. MenTeR is designed with interactive checkpoints where human designers can intervene as needed.}
    \label{fig:figure_1}
    \vspace{-10pt}
    
\end{figure*}

To address these challenges, we propose MenTeR: A fully-automated multi-agent workflow for end-to-end RF/Analog circuits netlist design, illustrated in Figure \ref{fig:figure_1}. MenTeR is designed to facilitate analog designers with specification reasoning, document search, testbench generation, and schematic design. Our contributions are:
\begin{enumerate}
    \item To the best of our knowledge, MenTeR is the first fully-automated multi-agent framework for end-to-end RF/Analog design using LLM, which can generate analog circuit netlists of a single-block, multi-blocks, and ultimately system-level through a scalable approach;
    \item A practical study of LLMs' reasoning capabilities enhancement with Diagram-Aware Retrieval-Augmented Generation (DA-RAG) and self-referential techniques, which also increases LLMs' capabilities in test bench writing, leading to minimal human engineer intervention;
    \item We introduce a Chain-of-Stage (CoS) reasoning agent for RF/Analog design that integrates domain-specific knowledge with the Chain-of-Thought (CoT) technique, enhancing the reasoning capabilities of LLMs without requiring extensive training data.
\end{enumerate}

\section{Background}

\subsection{Enhancing Reasoning Capabilities of LLMs}

In this section, we review key techniques for enhancing LLMs' reasoning capabilities, which inform our approach to improving LLMs' RF/Analog design performance. 

\textbf{Prompt Engineering.} 
Deng et al. \cite{b6} proposed a "Rephrase and Respond" (RaR) method that enables language models to rephrase input questions for better clarity, significantly improving models' performance by resolving ambiguities in human-LLM communication. Wei et al. \cite{b7} introduced Chain-of-Thought (CoT) prompting, where models were guided to generate step-by-step reasoning processes before producing final answers, improving their performance on complex reasoning tasks. Based on this approach, Yao et al. \cite{b8} proposed Tree of Thoughts (ToT) prompting, improving performance over CoT prompting in complex reasoning tasks by allowing models to explore and evaluate multiple reasoning paths. 

These  prompting techniques provide efficient mechanisms for us to evaluate and enhance LLMs' capabilities in analog design, where structured reasoning and alternative solution explorations are particularly valuable. 

\textbf{Retrieval-Augmented Generation.} 
Retrieval-Augmented Generation (RAG) proposed by Lewis et al. \cite{b9} has been a widely used technique to increase the ability of LLMs in knowledge-intensive language tasks. By combining a pre-trained seq2seq model with a dense vector retrieval system over text-based documents,  it enables knowledge updates without model retraining. In analog design, \cite{b4} and \cite{b10} have used the basic RAG technique with analog design documents to enhance LLMs' capabilities in a complex design task. 

\textbf{Fine Tuning.}
Moreover, aligning language models with human preferences has proven to be effective in enhancing reasoning capabilities. Ouyang et al. \cite{b11} demonstrated that fine-tuning language models with Reinforcement Learning from Human Feedback (RLHF) significantly improves model alignment with user intentions. However, RLHF is a complex and often unstable procedure that involves two main steps: 
\begin{enumerate}
    \item Fitting a reward model to reflect human preferences.
    \item Fine-tuning a language model with reinforcement learning to maximize this estimated reward.
\end{enumerate}
To address these challenges, Rafailov et al. \cite{b12} introduced Direct Preference Optimization (DPO), a method that bypasses explicit reward modeling and reinforcement learning by directly training language models from human preferences through a simple classification loss. Building on this approach, Meng et al. \cite{b13} proposed SimPO, a simplified preference optimization approach that improved DPO by using a reference-free reward formulation based on the average logarithmic probability of a sequence. Most recently, DeepSeek \cite{b14} demonstrated that reasoning capabilities can be significantly enhanced by large-scale reinforcement learning on high-quality Chain-of-Thought (CoT) data.

When we take a closer look at these fine-tuning methodologies, it becomes apparent that effective RF/Analog design fine-tuning requires either sophisticated reward modeling tailored to circuit designers' preferences or the development of a high-quality CoT dataset specific to the designs. Both approaches represent substantial long-term commitments. In this paper, we pay attention to the latter direction. By integrating our multi-agent workflow into daily design processes, we aim to not only solve fundamental RF/Analog circuit problems but also collect the reasoning paths employed in analog circuit design systematically. This helps us prepare the necessary steps for building a domain-adapted reasoning model to assist with more complex circuit designs.



\begin{figure}[H]
    \centerline{\includegraphics[width=\columnwidth, height=8cm]{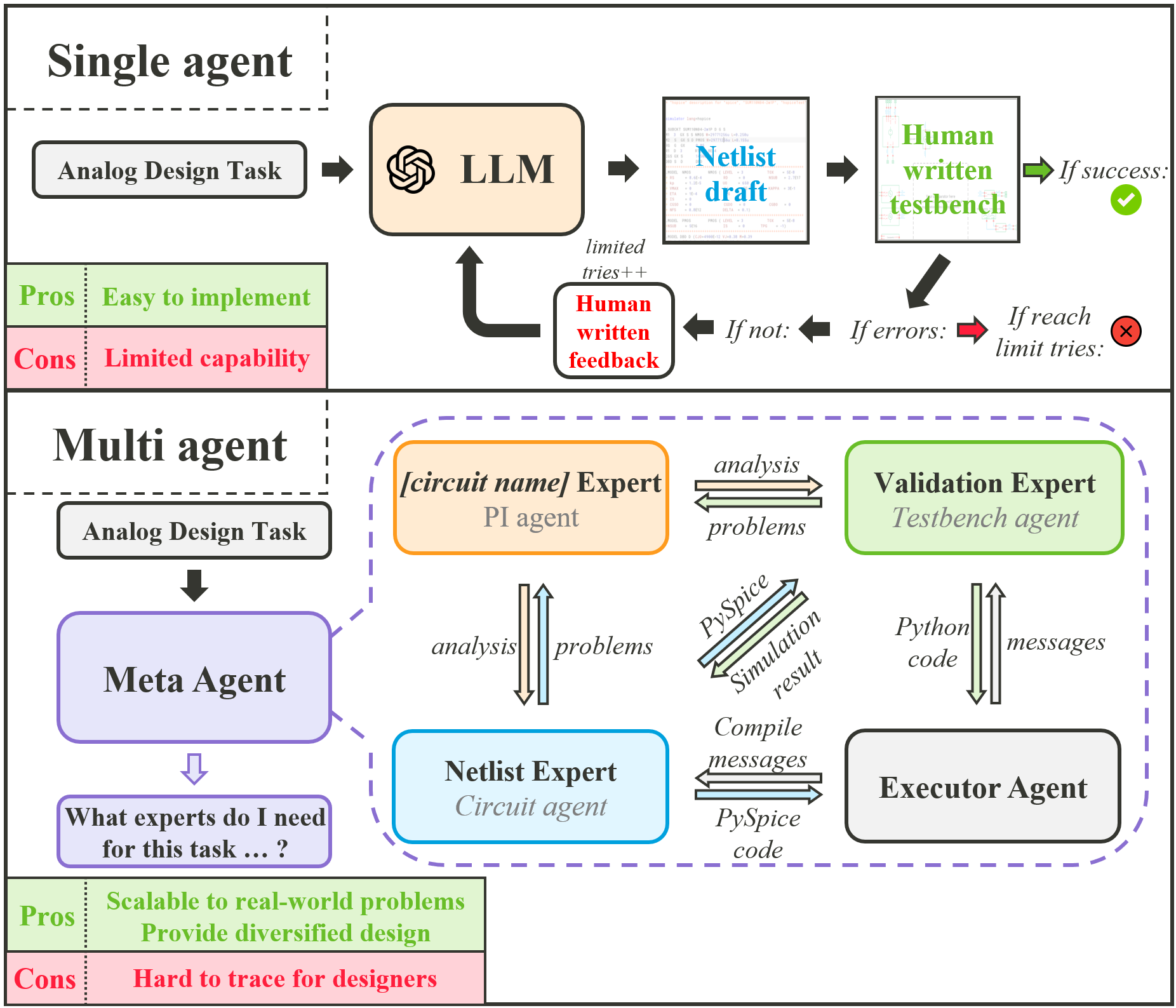}}
    \vspace{-4pt}
    \caption{Comparison between single-agent and simplified multi-agent frameworks.}
    \label{fig:figure_2}
\end{figure}

\subsection{The Necessity of a Multi-Agent Workflow}

Here, we investigate the importance of a multi-agent workflow to achieve the aforementioned goal. Wu et al. \cite{b15} developed AutoGen, an open-source framework that facilitated multi-agent conversations between customizable LLM-powered agents, human inputs, and tools to tackle complex tasks across mathematics, coding, and decision-making domains. Building on this concept, Song et al. \cite{b16} proposed Captain Agent, an adaptive framework that dynamically assembled and managed teams throughout the task-solving process, providing a valuable reference for formulating our multi-agent workflow. Based on these insights, we have implemented a simplified multi-agent framework and compared it with a single-agent approach, as illustrated in Figure \ref{fig:figure_2}.

While a single-agent approach offers implementation simplicity for analog design tasks, it inevitably encounters bottlenecks limited by the capabilities of the underlying Large Language Model. As we attempted to solve complex design challenges using LLM-based methodologies, our proposed multi-agent system demonstrated significant benefits. 

In our simplified multi-agent framework, we implemented a Meta Agent serving as the central orchestrator, which analyzes incoming design tasks to determine which specialized domain experts are required. Upon receiving an analog design specification, the Meta Agent decomposes complex problems into manageable sub-tasks, delegates these components to appropriate domain specialist agents, and oversees the entire workflow.

With this simplified multi-agent approach, our method already outperformed single-agent work such as \cite{b3} (see Table \ref{tab:tab_1} for more details). However, the structure of this framework can be further optimized to integrate into contemporary design flows. The detail of our proposed MenTeR framework will be interpreted in detail in Section \ref{sec:sec_3}.


\section{Overview of MenTeR}
\label{sec:sec_3}

\begin{figure}[H]
    \centerline{\includegraphics[width=\columnwidth, height=9cm]{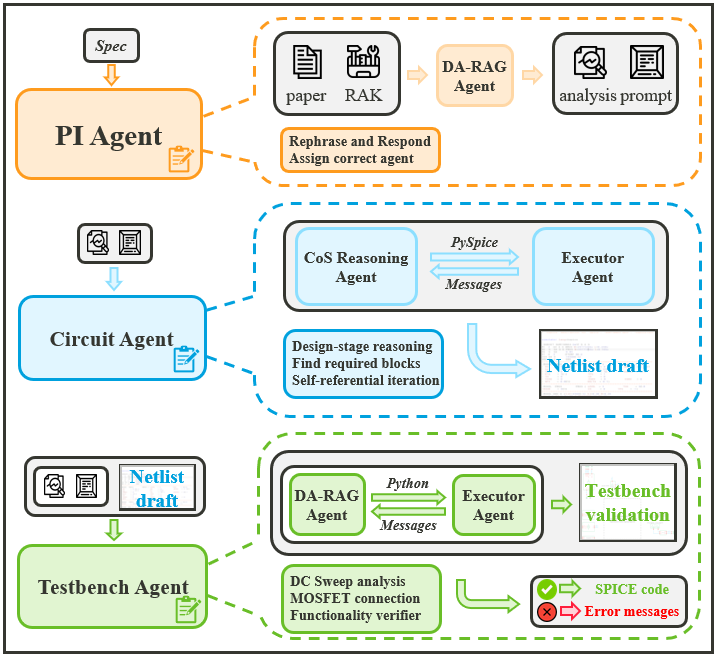}}
    \vspace{-4pt}
    \caption{Overview of MenTeR agents.}
    \label{fig:figure_3}
\end{figure}

In this section, we introduce the proposed MenTeR framework, which decomposes an analog design task into multiple stages, including specification understanding, document search, circuit netlist design, and test bench generation. The overview of MenTeR is illustrated in Figure \ref{fig:figure_1}, and the flow of each agent is illustrated in Figure \ref{fig:figure_3}.



\subsection{PI Agent}\label{AA}
PI agent is designed to play the role of the "Primary Investigator," who interacts with the users, manages the workflow, and assigns tasks to corresponding agents. Given an analog design task, the PI agent should leverage the DA-RAG agent to process design requirements using multiple knowledge sources: technical papers (e.g., \cite{b18, b20}), Rapid Adoption Kit (RAK), and textbook chapters (e.g., \cite{b17}). 

After retrieving the knowledge, PI agent defines the tasks and assigns them to different agents respectively. This paradigm transforms abstract user requirements into well-defined specifications, which bridges the semantic gap between human design intent and formal circuit requirements.

\subsection{Circuit Agent}\label{AA}
Upon receiving processed requirements from the PI Agent, the Circuit Agent (CA) functions as a "senior analog designer," generating the actual circuit design in PySpice\footnote{https://github.com/PySpice-org/PySpice} format. The CA's internal architecture consists of two primary components: the Chain-of-Stage (CoS) agent and the Executor agent. The CoS agent manages design stage reasoning, identifies familiar circuit blocks, and conducts self-referential iteration for design refinement. 

When tasked with a design challenge, the Circuit Agent first determines the necessary design stages required for the desired circuit implementation. For familiar design blocks, it can consult domain experts in the Circuit Think Tank to leverage established design patterns. After formulating the initial design, the Circuit Agent produces a netlist draft which is then passed to the Executor agent. This secondary agent performs essential validation checks including syntax verification and other critical error detection processes. Any errors identified during this preliminary validation are routed back to the CoS agent for targeted refinement. 

Once the netlist passes these basic verification checks, the validated draft is forwarded to the Testbench Agent for comprehensive simulation and testing.

\subsection{Testbench Agent}\label{AA}

By leveraging DA-RAG (refers to Section \ref{sec:sec_3_e}), the TestBench Agent (TBA) defines appropriate simulations required for the specific circuit, implementing three specialized validation components: DC sweep checker, MOSFET (and other primitive elements such as resistors) connection checker, and functionality verifier. 
TBA generates corresponding Python code to validate the PySpice implementation. In the future, more advanced check agents such as steady-state large-signal operating point check can be included to facilitate system-level design.

Similar to the Circuit Agent's workflow, the testbench code is passed to an Executor agent to verify syntax correctness and ensure proper simulation. Upon completion of the validation process, it either confirms successful design verification or produces detailed error messages that are routed back to the Circuit Agent for targeted design refinement.

\subsection{Circuit Think Tank (CTT)}\label{AA}
As we deploy MenTeR in real-world applications, we anticipate encountering increasingly diverse circuits that share fundamental design concepts. To leverage these similarities and reduce complexity, we established the Circuit Think Tank—a knowledge repository of specialized circuit expertise. 

After a circuit design is successfully completed and validated by the Testbench Agent, we capture its critical attributes including circuit name, specifications, reasoning stages, netlist implementation, and relevant interaction history. This repository serves dual purposes: it facilitates accelerated reasoning for new analog design tasks by providing access to established design patterns, and it constitutes a valuable dataset for future domain-adapted fine-tuning of our agents.

\begin{figure}[H]
    \centerline{\includegraphics[width=\columnwidth]{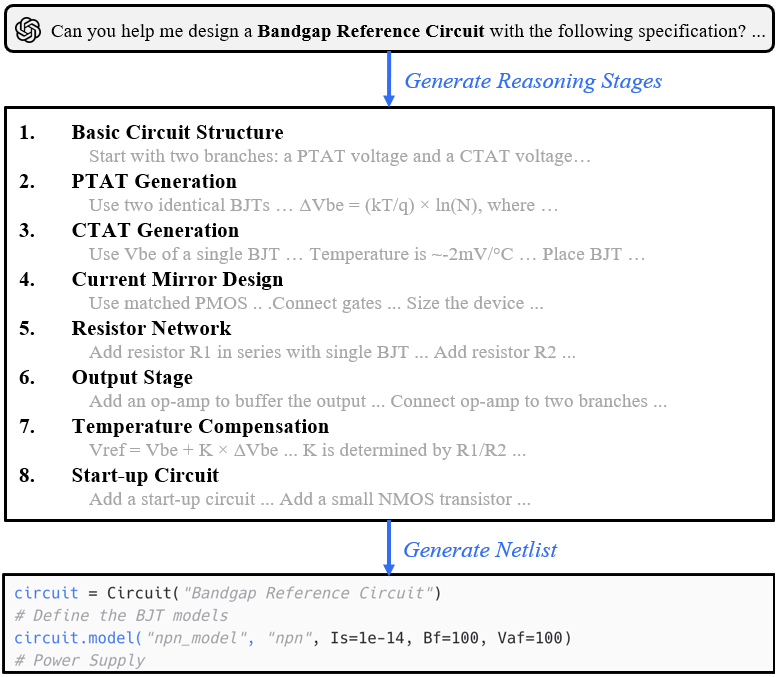}}
    \vspace{-4pt}
    \caption{CoS reasoning stages for the BGR circuit.}
    \label{fig:figure_4}
\end{figure}

\subsection{Sub-Agents}\label{AA}
Moreover, we construct several sub-agents to equip techniques for each agent, including:

\subsubsection{\textbf{CoS Reasoning Agent}}
In contrast to conventional one-pass approaches, we propose a CoS reasoning agent to decompose the RF/Analog design problem into a sequence of specialized sub-tasks, or “stages.” Figure \ref{fig:figure_4} illustrates this methodology through the reasoning stages developed for a Bandgap Reference circuit. By treating each stage as an independent task that feeds its intermediate outputs into the subsequent stage, CoS ensures that critical design stages and constraints are addressed during the generation process.

For example, the first stage of our CoS reasoning agent leverages task-relevant information (e.g. from textbooks or prior designs know-how from agents in CTT) to establish the fundamental design requirements. The system then transitions into a parameter synthesis stage, where CoT will be used to follow this hierarchical system to propose initial component values. Specifically, each stage stores both its prompts and outputs as prior knowledge, allowing for iterative updates and cross-referencing between stages.

\subsubsection{\textbf{Diagram-Aware RAG (DA-RAG) Agent}}
\label{sec:sec_3_e}

\begin{figure}[H]
    \vspace{-4pt}
    \centerline{\includegraphics[width=\columnwidth]{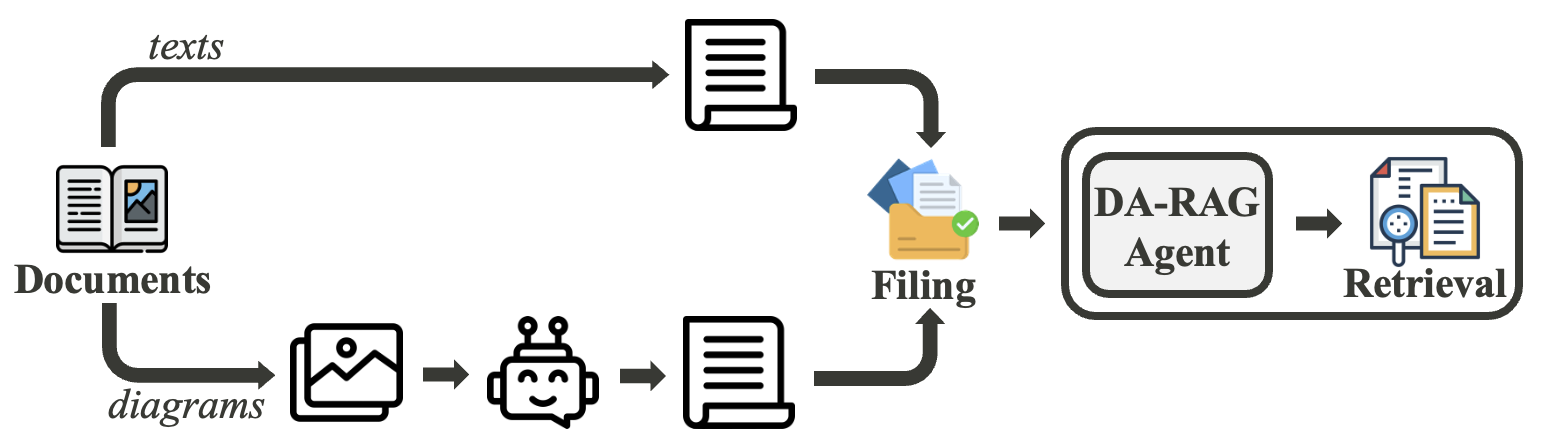}}
    \vspace{-4pt}
    \caption{Overview of Diagram-Aware RAG (DA-RAG).}
    \label{fig:figure_5}
\end{figure}

In RF/Analog design, diagrams commonly encapsulate substantial information—such as schematic topologies, performance curves, and intricate design annotations—that cannot be readily conveyed through textual means. To effectively extract such information, we propose a Diagram-Aware RAG (DA-RAG) wherein real-world circuit textbooks \cite{b18}, often filled with specialized domain expertise, are methodically converted into Markdown format \cite{b19}, as illustrated in Figure \ref{fig:figure_5}. Unlike conventional RAG-based frameworks that primarily process text, DA-RAG leverages a LLM to transform these diagrams into textual representations, thereby enabling downstream retrieval and capture design requirements comprehensively. 


\vspace{-8pt}
\subsubsection{\textbf{Executor Agent}}
Similar to a compiler, Executor Agent functions as a self-referential system that syntactically validates generated code and circuit descriptions, ensuring their correctness and executability before passing them to subsequent stages of the workflow.



\begin{table*}[htbp]
\centering
\small
\caption{Performance comparison (in percentage) on 24 analog design tasks and a BGR problem. The highest scores for the hard tasks and average performance are highlighted as \textbf{bold}.
 (Task~1--8: Easy, Task~9--13: Medium, Task~14--24 \& BGR: Hard)}

\label{tab:tab_1}
\begin{adjustbox}{max width=\textwidth}
\begin{tabular}{|c|rr|rr|rr|rr|rr|rr|rr|}
\hline
\multicolumn{1}{|c|}{Method} & 
  \multicolumn{6}{c|}{AnalogCoder (Single-Agent)} &
  \multicolumn{2}{c|}{\makecell{Multi-Agent \\ Workflow (ours)}} &
  \multicolumn{2}{c|}{\makecell{MenTeR w/o \\ DA-RAG (ours)}} &
  \multicolumn{2}{c|}{\makecell{MenTeR w/o \\ CoS (ours)}} &
  \multicolumn{2}{c|}{\textbf{\makecell{MenTeR \\ (ours)}}} \\
\cline{1-15}
\multicolumn{1}{|c|}{Model} &
  \multicolumn{2}{c|}{\makecell{DeepSeek R1 \\ Distill 32B}} &
  \multicolumn{2}{c|}{\makecell{GPT-4o}} &
  \multicolumn{2}{c|}{\makecell{GPT-o3-mini}} &
  \multicolumn{8}{c|}{\makecell{GPT-4o}} \\
\cline{1-15}

 \multicolumn{1}{|c|}{Task} & 
  \multicolumn{1}{c}{Pass@1} &\multicolumn{1}{c|}{Pass@5} &
  \multicolumn{1}{c}{Pass@1} &\multicolumn{1}{c|}{Pass@5} &
  \multicolumn{1}{c}{Pass@1} &\multicolumn{1}{c|}{Pass@5} &
  \multicolumn{1}{c}{Pass@1} &\multicolumn{1}{c|}{Pass@5} &
  \multicolumn{1}{c}{Pass@1} &\multicolumn{1}{c|}{Pass@5} &
  \multicolumn{1}{c}{Pass@1} &\multicolumn{1}{c|}{Pass@5} &
  \multicolumn{1}{c}{Pass@1} &\multicolumn{1}{c|}{Pass@5} \\
\hline
1   &   100.0 & 100.0 &  100.0 & 100.0 &   60.0 &  80.0 & 100.0 & 100.0 &  80.0 & 100.0 & 100.0 & 100.0 &          100.0 &          100.0\\
2   &    60.0 & 100.0 &  100.0 & 100.0 &   40.0 & 100.0 & 100.0 & 100.0 & 100.0 & 100.0 &  80.0 & 100.0 &          100.0 &          100.0\\
3   &    20.0 &  60.0 &  100.0 & 100.0 &   80.0 & 100.0 &  80.0 & 100.0 & 100.0 & 100.0 & 100.0 & 100.0 &          100.0 &          100.0\\
4   &    40.0 &  40.0 &   40.0 &  80.0 &   60.0 & 100.0 &  80.0 & 100.0 & 100.0 & 100.0 & 100.0 & 100.0 &          100.0 &          100.0\\
5   &    20.0 &  40.0 &    0.0 &  40.0 &   80.0 &  80.0 & 100.0 & 100.0 &  80.0 & 100.0 &  80.0 & 100.0 &          100.0 &          100.0\\
\hline
6   &   100.0 & 100.0 &  100.0 & 100.0 &   60.0 & 100.0 & 100.0 & 100.0 & 100.0 & 100.0 & 100.0 & 100.0 &          100.0 &          100.0\\
7   &    80.0 &  80.0 &   80.0 & 100.0 &   20.0 & 100.0 & 100.0 & 100.0 &  80.0 & 100.0 & 100.0 & 100.0 &           80.0 &           80.0\\
8   &    60.0 &  60.0 &  100.0 & 100.0 &   60.0 &  80.0 & 100.0 & 100.0 & 100.0 & 100.0 &  80.0 &  80.0 &          100.0 &          100.0\\
9   &    20.0 &  60.0 &   60.0 & 100.0 &   60.0 & 100.0 & 100.0 & 100.0 & 100.0 & 100.0 &  40.0 &  60.0 &          100.0 &          100.0\\
10  &     0.0 &  60.0 &   60.0 & 100.0 &   80.0 & 100.0 & 100.0 & 100.0 &  60.0 &  60.0 & 100.0 & 100.0 &          100.0 &          100.0\\
\hline
11  &     0.0 &   0.0 &    0.0 &  20.0 &   20.0 & 100.0 &  20.0 &  40.0 &  20.0 &  60.0 &  20.0 &  20.0 &           40.0 &           60.0\\
12  &     0.0 &   0.0 &    0.0 &   0.0 &   20.0 &  80.0 &   0.0 &   0.0 &  40.0 &  60.0 &  20.0 &  60.0 &           40.0 &           40.0\\
13  &     0.0 &   0.0 &   40.0 &  60.0 &   60.0 & 100.0 &  20.0 &  20.0 &   0.0 &   0.0 &  20.0 &  40.0 &           20.0 &           20.0\\
14  &     0.0 &   0.0 &    0.0 &  60.0 &    0.0 &  60.0 &  20.0 &  20.0 &  40.0 &  40.0 &  20.0 &  40.0 &  \textbf{40.0} &  \textbf{60.0}\\
15  &     0.0 &   0.0 &    0.0 &   0.0 &   20.0 &  20.0 &  20.0 &  20.0 &   0.0 &   0.0 &  20.0 &  40.0 &  \textbf{20.0} &  \textbf{80.0}\\
\hline
16  &    60.0 &  60.0 &   80.0 &  80.0 &   40.0 &  80.0 & 100.0 & 100.0 &  60.0 &  60.0 & 100.0 & 100.0 & \textbf{100.0} & \textbf{100.0}\\
17  &    60.0 &  60.0 &   20.0 &  20.0 &    0.0 &  40.0 &  60.0 &  60.0 & 100.0 & 100.0 & 100.0 & 100.0 & \textbf{100.0} & \textbf{100.0}\\
18  &    20.0 &  20.0 &   20.0 &  20.0 &   60.0 & 100.0 &  80.0 &  80.0 & 100.0 & 100.0 & 100.0 & 100.0 & \textbf{100.0} & \textbf{100.0}\\
19  &    60.0 &  60.0 &   60.0 &  60.0 &   60.0 & 100.0 &  60.0 &  60.0 & 100.0 & 100.0 & 100.0 & 100.0 & \textbf{100.0} & \textbf{100.0}\\
20  &    60.0 &  60.0 &   20.0 &  20.0 &   20.0 &  60.0 & 100.0 & 100.0 & 100.0 & 100.0 &  80.0 &  80.0 & \textbf{100.0} & \textbf{100.0}\\
\hline
21  &    40.0 &  40.0 &   60.0 &  60.0 &   40.0 &  80.0 & 100.0 & 100.0 & 100.0 & 100.0 & 100.0 & 100.0 & \textbf{100.0} & \textbf{100.0}\\
22  &    40.0 &  40.0 &   40.0 &  40.0 &   40.0 &  80.0 &  60.0 &  60.0 &  80.0 &  80.0 & 100.0 & 100.0 & \textbf{100.0} & \textbf{100.0}\\
23  &     0.0 &   0.0 &    5.0 &  60.0 &   60.0 &  80.0 & 100.0 & 100.0 & 100.0 & 100.0 &  20.0 &  60.0 & \textbf{100.0} & \textbf{100.0}\\
24  &     0.0 &   0.0 &   10.0 &  20.0 &   20.0 &  40.0 &   0.0 &  20.0 &  60.0 &  60.0 &   0.0 &  0.0 &  \textbf{80.0} & \textbf{100.0}\\
\hline
\textbf{Avg}  & 35.5 & 43.3 & 45.6 & 60.0 & 44.2 & 81.7 & 70.8 & 74.2 & 75.0 & 80.0 & 70.0 & 78.3 & \textbf{84.2} & \textbf{89.2}\\ 
\hline
\textbf{BGR} & 0.0 &  0.0 &  0.0 &  0.0 & 0.0 &  0.0 & 60.0 & 100.0 & 60.0 & 80.0 & 60.0 & 80.0 & \textbf{80.0} & \textbf{100.0}\\
\hline
\end{tabular}
\end{adjustbox}
\vspace{-10pt}
\end{table*}

\section{Experiment Results}

To evaluate the effectiveness of our framework, we compared MenTeR against several baseline approaches, including individual LLMs (single-agent) and multi-agent systems. We evaluated them on 24 standard analog design tasks from \cite{b3}, ranging from elementary designs to complex multi-block design challenges. Furthermore, we deployed our workflow to solve a CMOS Bandgap Reference circuit, a practical task encountered in real-world implementations, to validate MenTeR's performance in industrial applications.


\subsection{Experimental Setup}

We evaluated all methods by generating circuit netlists and checking whether the produced solution passed functional validation. Specifically, we considered the \emph{pass@k} metric, which measures the probability of obtaining at least one correct solution within \emph{k} attempts:
\begin{equation}
\text{pass@k} = 1 - \frac{\binom{n - c}{k}}{\binom{n}{k}}
\end{equation}
where:
\begin{itemize}
    \item $n$ is the total number of solutions generated,
    \item $c$ is the number of correct solutions,
    \item $k$ is the success threshold (number of attempts).
\end{itemize}
A circuit is deemed ``correct'' if it meets the specified performance criteria (e.g., gain, power consumption, linearity, bandwidth, etc.) To ensure consistency, all solutions generated were further inspected with basic electrical rule checks and circuit simulations, verifying that the netlists could be functionally simulated without errors.

\begin{figure}[H]
    \centerline{\includegraphics[width=\columnwidth]{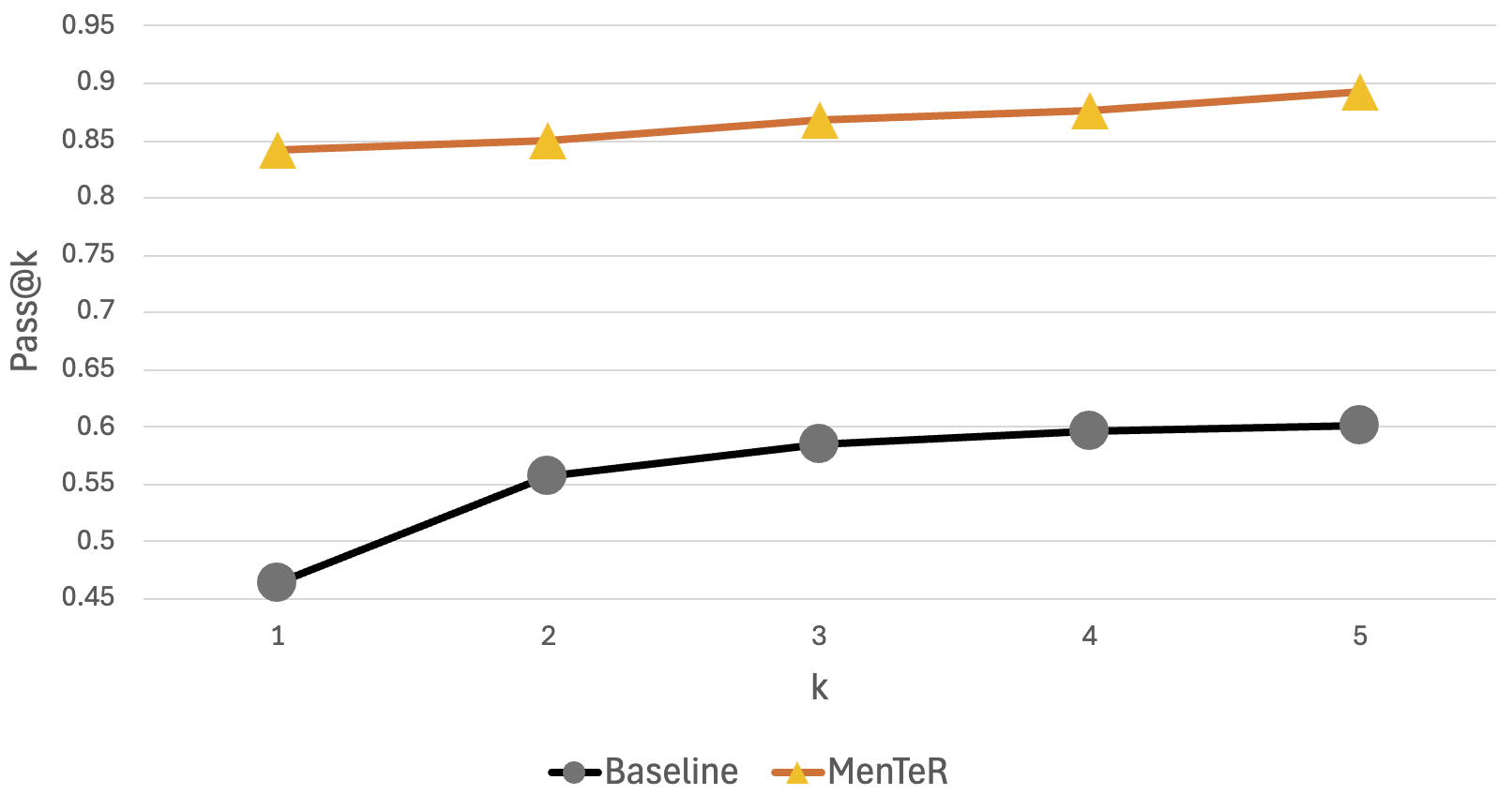}}
    \vspace{-8pt}
    \caption{\emph{Pass@k} comparison between single-agent baseline and MenTeR.}
    \vspace{-8pt}
    \label{fig:figure_6}
\end{figure}

\subsection{Comparison Results}

Table \ref{tab:tab_1} presents our experimental results, with testing limited to five attempts per task for cost effectiveness. 
We used GPT-4o as the backbone model for all our approaches to maintain cost efficiency and ensure fair comparison with single-agent methods. For baseline single-agent evaluations, we selected AnalogCoder \cite{b3} implemented with distilled model variants (e.g. OpenAI o3-mini \cite{b23}, DeepSeek-R1 distilled 32B \cite{b24}) rather than state-of-the-art reasoning models (e.g. OpenAI o1 \cite{b21}, DeepSeek-R1 671B \cite{b22}), considering the substantial inference costs associated with the latter.


    

\subsubsection{\textbf{Overall Performance}}
MenTeR demonstrates strong performance in various analog design tasks, achieving an average Pass@1 rate of 84.2\%. Specifically, MenTeR consistently solves basic circuits (e.g., Task~1--6) within a single attempt and outperforms other methods in many advanced tasks requiring multi-block circuit interplay. This significantly outperforms the single-agent baseline, even when the baseline is given multiple attempts (at least five tries per task), as illustrated in Figure \ref{fig:figure_6}.



Remarkably, MenTeR achieves a successful Pass@1 rate of at least 80\% and a perfect 100\% at Pass@5 rate for complex tasks (e.g., Task~16--24) such as Schmitt trigger, Voltage-Controlled Oscillator, and low-voltage BGR circuit. This indicates its potential to assist real-world complex analog design tasks.

\subsection{Ablation Studies}


To isolate the contribution of each MenTeR component, we conducted our ablation study to examine the contributions of MenTeR's key components, as demonstrated in Table \ref{tab:tab_2}.
\begin{itemize}
    \item \textbf{Without DA-RAG:} Replacing the DA-RAG agent inside \textit{PI Agent} and \textit{Testbench Agent} with a generic approach decreased the average pass@1 by about 9\%. This performance degradation highlighted LLM knowledge limitations when operating without specialized domain know-how. Without access to task-relevant documents, it failed to effectively address advanced analog design tasks that require specific circuit knowledge beyond the model's inherent capabilities.
    
    
    \item \textbf{Without CoS Reasoning:} Removing CoS logic left the multi-agent system with a more “flat” approach to design. Pass@1 dropped significantly (maximum by 14\%), especially on tasks with multiple nested constraints. Notably, we observed that the Pass@1 rate of MenTeR without CoS fell below the multi-agent baseline, suggesting that without structured reasoning guidance, the system becomes overwhelmed by retrieved document information and fails to prioritize relevant design knowledge.
    
\end{itemize}

\begin{table}[H]
\centering
\caption{Ablation Studies Results for MenTeR Components.}
\begin{tabular}{|l|c|c|}
\hline
\textbf{Method} & \textbf{Avg. Pass@1 (\%)} & \textbf{Avg. Pass@5 (\%)} \\
\hline
Multi-Agent Workflow & 70.8 & 74.2 \\
MenTeR w/o DA-RAG & 75.0 & 80.0 \\
MenTeR w/o CoS & 70.0 & 78.3 \\
Full MenTeR & \textbf{84.2} & \textbf{89.2} \\
\hline
\end{tabular}
\label{tab:tab_2}
\end{table}

These findings substantiate the necessity of both DA-RAG and CoS reasoning within the MenTeR framework, demonstrating how each component addresses specific limitations in LLM-based analog circuit design.

\section{Conclusion}

In this work, we propose MenTeR, a fully-automated multi-agent workflow for RF/Analog circuit design that scales effectively to system-level circuit design. By leveraging techniques including Diagram-Aware Retrieval-Augmented Generation (DA-RAG), Chain-of-Stage (CoS) reasoning, and self-referential mechanisms, MenTeR successfully addresses various fundamental circuit blocks without requiring human intervention or extensive fine-tuning. Specifically, MenTeR achieves significantly better performance when handling increasingly complex circuits compared to single-agent approaches, validating its potential for scaling up to system-level circuits and integration with contemporary design flows.

\textbf{Opportunities and Future work.} We acknowledge that MenTeR's capabilities remain highly correlated with the underlying large language models that power its agents. For more complex analog design tasks, domain-adapted reasoning models specifically fine-tuned for analog design could be a promising direction for future development. Throughout MenTeR's operation, we have deliberately structured the system to collect high-quality design data, with the goal of building a comprehensive analog design reasoning dataset. We expect this framework will facilitate the development of LLM-driven EDA tools for analog circuits, paving the way towards collaborative teamwork between human analog designers and an RF/Analog copilot.

\section*{Acknowledgement}
We would like to thank \href{mailto:aoowweenn@gmail.com}{Wei-Chen Chien} and \href{mailto:minchun.wu@alumni.psu.edu}{Min-Chun Wu} for their valuable consultation on the agent framework.
\begin{appendices}

\section{MenTeR Stability}

We evaluated MenTeR using different backbone models, as summarized in Table \ref{tab:menter_model_comparison}. MenTeR achieves at least a 77\% pass@1 rate and 85\% pass@5 rate across both models, demonstrating its stable performance. Notably, GPT-4o outperforms GPT-4.1, with higher average pass rates in both metrics. The slightly lower performance of GPT-4.1 may be attributed to its relative deficiency in Analog IC knowledge, which we discuss further in Appendix \ref{sec:aic_benchmark}.

\begin{table}[H]
\vspace{-4pt}
\centering
\caption{MenTeR with different backbone models.}
\begin{tabular}{|c|c|c|}
\hline
\textbf{MenTeR Model} & \textbf{Avg. Pass@1 (\%)} & \textbf{Avg. Pass@5 (\%)} \\
\hline
GPT-4o & \textbf{84.2} & \textbf{89.2} \\
GPT-4.1 & 77.5 & 85.8 \\
\hline
\end{tabular}
\label{tab:menter_model_comparison}
\vspace{-10pt}
\end{table}

\section{Comparison with Real-world Design}

For the Phase-Locked Loop (PLL) task, we compared AI-generated results with a human-designed reference. Our analysis revealed discrepancies between circuits that passed the benchmark validation and those that were functionally correct. 

\begin{figure}[H]
    \centerline{\includegraphics[width=\columnwidth]{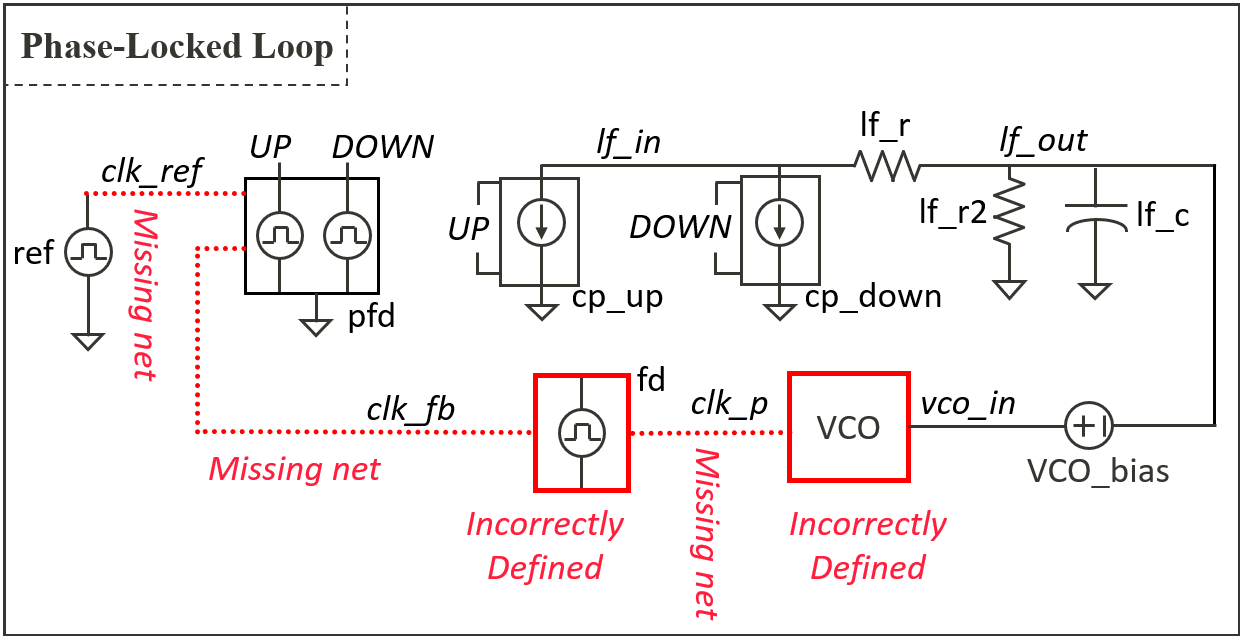}}
    \vspace{-4pt}
    \caption{Schematic of an AI generated Phase-Locked Loop (PLL).}
    \label{fig:figure_7}
\end{figure}

For instance, Figure \ref{fig:figure_7} shows a PLL implementation that was deemed correct by the benchmark criteria but contained errors upon schematic inspection. These errors primarily fell into two categories: incorrectly defined subcircuits (Frequency Detector (FD), Voltage Controlled Oscillator (VCO)), and missing wire connections (VCO to FD, FD to Phase Frequency Detector (PFD), and PFD to the reference frequency source). These issues stem from the inherent complexity of the PLL as a hierarchical system-level design. Notably, while individual subcircuits successfully passed DC sweep simulations in isolation, the integrated system as a whole failed to function correctly. This highlights the challenges of fully automating the generation of complex system-level designs using LLM-based RF/Analog design tools.

To further deploy MenTeR into real-world design environments, maintaining a "human-in-the-loop" approach remains essential. Such human oversight is particularly critical for complex system-level designs like PLL, where subtle integration errors may escape automated validation. Nevertheless, as the analog design reasoning capabilities of LLMs continue to advance, we can reasonably expect to reduce human intervention significantly, allowing designers to focus their expertise on system-level verification and optimization rather than routine circuit implementation tasks.

\vspace{-10pt}
\section{Engineering Challenges}

In our experiment, we encountered several engineering challenges that impacted efficiency and resource utilization. In this section, we share these challenges and their corresponding solutions to assist future LLM-based analog design research. 


\textbf{Token Limits.} 
The primary issue we encountered when testing on \cite{b3} was exceeding the token limits, which led to frequent crashes and interfered with our experimental progress. Upon investigation, we identified that the root cause stemmed from inefficient chat history parsing techniques that included the whole conversation texts in each iteration, which occupied the context window rapidly. To resolve this limitation, we modified our approach to truncate the chat history in each agent, retaining only the initial prompt and the last message. This optimization effectively reduced token consumption and eliminated system crashes. 


In addition, we acknowledge the concerns raised by reviewers regarding potential information loss when adopting context truncation techniques. To address this issue, we believe that approaches similar to the memory management system proposed by \cite{b26} represent promising avenues to solve the "lost-in-the-middle" issue.


\textbf{LLM Instability}.
Another substantial obstacle encountered was the instability introduced by iterative updates to the underlying large language model. Although no official statement has confirmed this phenomenon, there exists a widely discussed concern within the research community regarding sudden shifts in model behavior and performance after platform updates. Our own observations were consistent with these anecdotal reports; at various points in the experimental process, the model’s generative consistency and accuracy appeared to deteriorate following version changes and seed randomness. While further investigation is required to systematically verify the extent of this issue, the unanticipated variability in model output poses significant challenges for maintaining reproducibility and fine-tuning results in LLM-based analog design research. 

\textbf{Execution Errors.} 
We experienced a notably high rate of execution errors when reproducing single-agent results. This issue significantly interrupted the workflow as the generated code was often "unrunnable," preventing it from reaching the testbench validation stage. To mitigate this issue, we incorporated a self-referential technique into both multi-agent and MenTeR, ensuring all code is executable before submission. This technique substantially decreased execution error rates, which led to the stability of our methodologies (as illustrated in Figure \ref{fig:figure_6}), improving deployment efficiency of the overall system. 

\textbf{Repetitive Error Loops.} 
We observed that certain errors appeared repeatedly during our experimental iterations; LLMs continuously encountering the same issues and becoming stuck, even after receiving specific instructions to avoid them. This cycle led to task failures and resulted in substantial resource waste. We determined this occurred due to inherent limitations of LLMs. Consequently, we implemented an early-detection mechanism to identify such recurring errors, automatically terminating the chat when these patterns are detected. In future work, "preventing LLMs from repeating the same mistakes" represents an important research direction that needs further exploration.

\textbf{Token Efficiency.} 
Lastly, we compare the token efficiency of the single-agent approach and MenTeR, both utilizing GPT-4o as the backbone model. In this context, prompt tokens refer to the input tokens provided to the model, while completion tokens denote the output tokens generated by the model. As shown in Table \ref{tab: token-comparison}, MenTeR consumes more tokens for the relatively simple Task 1. However, for the more challenging Task 24, the single-agent method exhibits a substantial increase in token usage, yet still fails to solve the task. 

This highlights that while the single-agent approach remains effective for straightforward tasks, its capability diminish as task complexity increases. Balancing token efficiency and problem-solving capability, particularly in RF/Analog design, remains an important direction for future research.

\begin{table}[h]
    \centering
    \caption{Token Comparison between Single-Agent and MenTeR. (* means fail)}
    \label{tab: token-comparison}
    \begin{tabular}{|c|rr|rr|}
        \hline
        \multicolumn{1}{|l|}{} &
          \multicolumn{2}{c|}{\makecell{\textbf{Single-Agent}}} &
          \multicolumn{2}{c|}{\makecell{\textbf{MenTeR}}} \\
        \hline
        Token & Prompt & Completion & Prompt & Completion\\
        \hline
        Task 1 (Easy) & 1330 & 269 & 53991 & 2938 \\
        \hline
        Task 24 (Hard) & 98173 & 43310* & 39038 & 2428 \\
        \hline
    \end{tabular}
\end{table}

\section{Alternative Schematic Exploration}
\label{sec:app_alt_schematic}
During our experiments, we noticed that the same problem specification can yield multiple valid solutions with slightly different transistor arrangements or bias schemes. For instance, in the tasks labeled as “Problem 20,” the system generated two distinct netlists (20-0 vs.~20-1), all claiming to meet similar specifications. 

We hypothesize that this behavior arises because the given problem constraints are not sufficiently strict to enforce a single unique topology or bias point. Despite leading to multiple functional solutions, such variability also highlights the LLM’s capacity to explore alternative schematics. Below, we include two representative netlists, which aim to implement a simple op-amp-based adder.

\subsection{Problem 20: Op-Amp Adder}

\lstdefinelanguage{SPICE}{
  morekeywords={.title,.subckt,.ends,.model,.include,.lib},
  morecomment=[l]{*},      
  sensitive=false
}

\lstset{
  language=SPICE,
  basicstyle=\scriptsize\ttfamily       
  keywordstyle=\color{blue}\bfseries,   
  commentstyle=\color{teal}\itshape,    
  numbers=left,                         
  numbersep=6pt,                        
  breaklines=true,                      
  frame=single,                         
  xleftmargin=1em,                      
  xrightmargin=1em
}

\textbf{Netlist \#20-0}
\begin{lstlisting}[basicstyle=\footnotesize\ttfamily,breaklines=true,columns=flexible,label={lst:opamp_adder}]
.title Opamp Adder
.subckt SingleStageOpamp Vinp Vinn Vout
Vdd Vdd 0 5.0
Vbias Vbias 0 1.5
M1 Voutp Vinp Source3 Source3 nmos_model l=1e-06 w=5e-05
M2 Vout Vinn Source3 Source3 nmos_model l=1e-06 w=5e-05
M3 Source3 Vbias 0 0 nmos_model l=1e-06 w=0.0001
M4 Voutp Voutp Vdd Vdd pmos_model l=1e-06 w=0.0001
M5 Vout Voutp Vdd Vdd pmos_model l=1e-06 w=0.0001
.model nmos_model nmos (kp=0.0001 level=1 vto=0.5)
.model pmos_model pmos (kp=5e-05 level=1 vto=-0.5)
.ends SingleStageOpamp

Vbias V_bias 0 1.79V
Vref Vref 0 1.79V
Vin1 Vin1 0 1.79V
Vin2 Vin2 0 1.915V
X1 V_bias Vinn Vout SingleStageOpamp
R1 Vin1 Vinn 20kOhm
R2 Vin2 Vinn 20kOhm
Rf Vout Vinn 20kOhm
Rref Vref Vinn 20kOhm
\end{lstlisting}

\textbf{Netlist \#20-1}
\begin{lstlisting}[basicstyle=\footnotesize\ttfamily,breaklines=true,columns=flexible]
.title Opamp Adder: Vout = -(Vin1+Vin2)
.subckt SingleStageOpamp Vinp Vinn Vout
Vdd Vdd 0 5.0
Vbias Vbias 0 1.5
M1 Voutp Vinp Source3 Source3 nmos_model l=1e-06 w=5e-05
M2 Vout Vinn Source3 Source3 nmos_model l=1e-06 w=5e-05
M3 Source3 Vbias 0 0 nmos_model l=1e-06 w=0.0001
M4 Voutp Voutp Vdd Vdd pmos_model l=1e-06 w=0.0001
M5 Vout Voutp Vdd Vdd pmos_model l=1e-06 w=0.0001
.model nmos_model nmos (kp=0.0001 level=1 vto=0.5)
.model pmos_model pmos (kp=5e-05 level=1 vto=-0.5)
.ends SingleStageOpamp

X1 bias inv Vout SingleStageOpamp
Vin1 Vin1 0 0V
Vin2 Vin2 0 0V
Vbias_source bias 0 1.79V
R1 Vin1 inv 10kOhm
R2 Vin2 inv 10kOhm
Rf Vout inv 10kOhm
Voffset_source voffset 0 3.58V
Roffset voffset inv 3.3333333333333335kOhm
\end{lstlisting}

Although each netlist relies on similar single-stage operational amplifier blocks, the resistor arrangements differ. These differences stem from multiple ways to achieve the same functional requirement (i.e., summing two input signals).

Overall, our findings indicate that when problem constraints are loosely defined, LLMs can explore multiple design topologies. While each solution may meet the functional specs at a high level, additional constraints (e.g. output swing requirements, matching conditions, noise margins) would be needed to converge on more uniform design outcomes.

\section{Analog IC Knowledge}

\label{sec:aic_benchmark}

In order to investigate the root causes of differing performance among LLM-based analog design approaches, we developed the Analog IC Benchmark (AICB) \cite{b25}. This dataset consists of \textbf{300 multiple-choice questions} derived from a standard analog IC (AIC) textbook, with each question providing \textbf{four possible answers}.

\subsection{Dataset Introduction}
To illustrate the format of AICB, we can consider the following example question:

\begin{quote}
\noindent\textbf{Question:} What is the primary function of a sample-and-hold circuit in an ADC?
\begin{enumerate}[label=(\Alph*)]
    \item To amplify the input signal
    \item To convert the analog signal to digital
    \item To hold the input signal constant in conversion
    \item To filter the input signal
\end{enumerate}
\end{quote}

The correct answer is \textbf{C}, reflecting a key concept in data conversion systems.
\subsection{Performance Results}

Table~\ref{tab: AICB-performance} summarizes the accuracies achieved by these models on the AICB dataset. As indicated, GPT-o3-mini attained the highest accuracy, closely followed by GPT-4o, while DeepSeek R1 Distill 32B showed noticeably lower performance.

\begin{table}[h]
    \centering
    \caption{Performance of LLMs on the AICB Dataset}
    \label{tab: AICB-performance}
    \begin{tabular}{|l|c|}
        \hline
        \textbf{Model} & \textbf{Accuracy (\%)} \\
        \hline
        DeepSeek R1 Distill 32B & 75.7 \\
        \hline
        GPT-4o & 85.0 \\
        \hline
        GPT-o3-mini & 91.3 \\
        \hline
        GPT-4.1 & 81.1 \\
        \hline
    \end{tabular}
\end{table}

\subsection{Key Observations}


\textbf{Instruction-Following Challenges in DeepSeek R1 Distill 32B.} We identify that DeepSeek R1 Distill 32B generates inconsistent output formatting (e.g., generating \texttt{**Answer**: A} or \texttt{\#\#\# Answer A} instead of the prescribed tag format), which contributes to the poor performance of the model within an agent-based system, as such systems often rely on strict text formats for downstream task coordination. Consequently, DeepSeek R1 Distill 32B's difficulties in adhering to instructions partially explain its poor results in Table \ref{tab:tab_1} when paired with the benchmark \cite{b3}.

\subsection{Limitations and Future Directions}

While AICB provides a preliminary means of assessing LLM proficiency in analog IC domains, it does not fully capture the complexities encountered in real-world circuit design. As a result, we plan to expand AICB by incorporating more challenging and open-ended questions, aligned with the multifaceted nature of industrial analog design tasks. Such an enhanced benchmark would offer deeper insights into each model’s capabilities and limitations, facilitating more robust research and development of LLM-based analog design tools.

\end{appendices}


\newpage
\begin{thebibliography}{00}
\bibitem{b1} M. Liu, T. Ene, R. Kirby, C. Cheng, N. Pinckney, R. Liang, J. Alben, H. Anand, S. Banerjee, I. Bayraktaroglu, B. Bhaskaran, B. Catanzaro, A. Chaudhuri, S. Clay, B. Dally, L. Dang, P. Deshpande, S. Dhodhi, S. Halepete, E. Hill, J. Hu, S. Jain, B. Khailany, K. Kunal, X. Li, H. Liu, S. Oberman, S. Omar, S. Pratty, A. Sarkar, Z. Shao, H. Sun, P. P. Suthar, V. Tej, K. Xu, and H. Ren, “Chipnemo: Domain-adapted llms for chip design,” 2023.
\bibitem{b2} J. Blocklove, S. Garg, R. Karri, and H. Pearce, “Chip-chat: Challenges and opportunities in conversational hardware design,” arXiv preprint arXiv:2305.13243, 2023.
\bibitem{b3} Y. Lai, S. Lee, G. Chen, S. Poddar, M. Hu, D. Z. Pan, and P. Luo, “Analogcoder: Analog circuit design via training-free code generation,” arXiv preprint arXiv:2405.14918, 2024.
\bibitem{b4} C. Liu, W. Chen, A. Peng, Y. Du, L. Du, and J. Yang, “Ampagent: An llm-based multi-agent system for multi-stage amplifier schematic design from literature for process and performance porting,” arXiv preprint arXiv:2409.14739, 2024.
\bibitem{b5} Y. Shi, Z. Tao, Y. Gao, T. Zhou, C. Chang, T. Wang, B. Chen, G. Zhang, A. Liu, Z. Yu, T. Lin, L. He, "AMSnet-KG: A Netlist Dataset for LLM-based AMS Circuit Auto-Design Using Knowledge Graph RAG", arXiv preprint arXiv:2411.13560, 2024.
\bibitem{b6} Y. Deng, W. Zhang, Z. Chen, and Q. Gu, "Rephrase and respond: Let large language models ask better questions for themselves" arXiv preprint arXiv:2311.04205
\bibitem{b7} J. Wei, X. Wang, D. Schuurmans, M. Bosma, E. Chi, Q. Le, and D. Zhou. "Chain of thought prompting elicits reasoning in large language models." arXiv preprint arXiv:2201.11903
\bibitem{b8} S. Yao, D. Yu, J. Zhao, I. Shafran, T. L. Griffiths, Y. Cao, and K. Narasimhan, “Tree of thoughts: Deliberate problem solving with large language models,” CoRR, vol. abs/2305.10601, 2023.
\bibitem{b9} P. Lewis, E. Perez, A. Piktus, F. Petroni, V. Karpukhin, N. Goyal, H. K¨ uttler, M. Lewis, W.-t. Yih, T. Rockt¨aschel et al., “Retrieval augmented generation for knowledge-intensive nlp tasks,” Advances in Neural Information Processing Systems, vol. 33, pp. 9459–9474, 2020.
\bibitem{b10} B. Liu et al., "LayoutCopilot: An LLM-Powered Multi-Agent Collaborative Framework for Interactive Analog Layout Design," in IEEE Transactions on Computer-Aided Design of Integrated Circuits and Systems, doi: 10.1109/TCAD.2025.3529805.
\bibitem{b11} L. Ouyang, J. Wu, X. Jiang, D. Almeida, C. Wainwright, P. Mishkin, C. Zhang, S. Agarwal, K. Slama, A. Ray, J. Schulman, J. Hilton, F. Kelton, L. Miller, M. Simens, A. Askell, P.Welinder, P. F. Christiano, J. Leike, and R. Lowe. Training language models to follow instructions with human feedback. In S. Koyejo, S. Mohamed, A. Agarwal, D. Belgrave, K. Cho, and A. Oh, editors, Advances in Neural Information Processing Systems, volume 35, pages 27730–27744. Curran Associates, Inc., 2022. 
\bibitem{b12} R. Rafailov, A. Sharma, E. Mitchell, S. Ermon, C. D. Manning, and C. Finn. Direct preference optimization: Your language model is secretly a reward model. In NeurIPS, 2023.
\bibitem{b13} Y. Meng, M. Xia, and D. Chen, SimPO: Simple Preference Optimization with a Reference-Free Reward. In NeurIPS, 2024.
\bibitem{b14} D. Guo, D. Yang, H. Zhang, J. Song, R. Zhang, R. Xu, Q. Zhu, S. Ma, P. Wang, X. Bi, et al. 2025. Deepseek-r1: Incentivizing reasoning capability in llms via reinforcement learning. arXiv preprint arXiv:2501.12948.
\bibitem{b15} Q. Wu, G. Bansal, J. Zhang, Y. Wu, B. Li, E. Zhu, L. Jiang, X. Zhang, S. Zhang, J. Liu, et al. 2024. Autogen: Enabling next-gen LLM applications via multi-agent conversation. In ICLR 2024 Workshop on Large Language Model (LLM) Agents.
\bibitem{b16} L. Song, J. Liu, J. Zhang, S. Zhang, A. Luo, S. Wang, Q. Wu, and C. Wang. Adaptive in-conversation team building for language model agents. arXiv preprint arXiv:2405.19425, 2024.
\bibitem{b17} B. Razavi. 2017. Design of Analog CMOS Integrated Circuits (2nd. ed.). McGraw-Hill, Inc., USA.
\bibitem{b18} B. Razavi, "The Design of a Low-Voltage Bandgap Reference [The Analog Mind]," in IEEE Solid-State Circuits Magazine, vol. 13, no. 3, pp. 6-16, Summer 2021, doi: 10.1109/MSSC.2021.3088963.
\bibitem{b19} Vik Paruchuri, Marker, 2023, GitHub, \url{https://github.com/VikParuchuri/marker}
\bibitem{b20} I. M. Filanovsky and H. Baltes, "CMOS Schmitt trigger design," in IEEE Transactions on Circuits and Systems I: Fundamental Theory and Applications, vol. 41, no. 1, pp. 46-49, Jan. 1994, doi: 10.1109/81.260219.
\bibitem{b21} OpenAI o1, https://openai.com/o1/
\bibitem{b22} DeepSeek R1, https://huggingface.co/deepseek-ai/DeepSeek-R1
\bibitem{b23} OpenAI o3-mini, https://openai.com/index/openai-o3-mini/
\bibitem{b24} DeepSeek R1 Distilled Qwen 32B, https://huggingface.co/deepseek-ai/DeepSeek-R1-Distill-Qwen-32B/
\bibitem{b25} AICB: Analog Integrated Circuit Benchmark, \url{https://github.com/treeleaves30760/AICB-Analog-Integrated-Circuit-Benchmark/}
\bibitem{b26} C. Packer, S. Wooder, K. Lin, et al. MemGPT: Towards LLMs as Operating Systems. arXiv preprint arXiv:2310.08560, 2023. 




\end{thebibliography}
\end{document}